%% file: main.tex
\documentclass[10pt,twocolumn,letterpaper]{article}

\usepackage{cvpr}              

\usepackage{graphicx}
\usepackage{amsmath}
\usepackage{amssymb}
\usepackage{booktabs}
\usepackage{times}
\usepackage{latexsym}
\usepackage{placeins}
\usepackage{booktabs}
\usepackage{multirow}
\usepackage{comment}
\usepackage{pifont}

\usepackage[pagebackref,breaklinks,colorlinks]{hyperref}

\usepackage[capitalize]{cleveref}
\crefname{section}{Sec.}{Secs.}
\Crefname{section}{Section}{Sections}
\Crefname{table}{Table}{Tables}
\crefname{table}{Tab.}{Tabs.}

\newcommand{\cotok}{Co-tokenization } 
\newcommand\blfootnote[1]{%
  \begingroup
  \renewcommand\thefootnote{}\footnote{#1}%
  \addtocounter{footnote}{-1}%
  \endgroup
}
\newcommand{\cmark}{\ding{51}}

\def\cvprPaperID{20} 
\def\confName{CVPR}
\def\confYear{2023}

\begin{document}

\title{Diversifying Joint Vision-Language Tokenization Learning}


\author{
Vardaan Pahuja\textsuperscript{1}$^{*}$,\quad
AJ Piergiovanni\textsuperscript{2},\quad
Anelia Angelova\textsuperscript{2}\\
\textsuperscript{1}The Ohio State University\quad\quad
\textsuperscript{2}Google DeepMind\\
{\small\tt pahuja.9@osu.edu, \{ajpiergi, anelia\}@google.com}\\
}
\maketitle

\begin{abstract}
   Building joint representations across images and text is an essential step for tasks such as Visual Question Answering and Video Question Answering.
   In this work, we find that the representations must not only jointly capture features from both modalities but should also be diverse for better generalization performance.
   To this end, we propose joint vision-language representation learning by diversifying the tokenization learning process, enabling tokens that are sufficiently disentangled from each other to be learned from both modalities.
    We observe that our approach outperforms the baseline models in a majority of settings and is competitive with state-of-the-art methods.\blfootnote{${}^*$Work done while at Google}
\end{abstract}

\input{intro}

\input{background}

\input{model}

\input{experiments}

\input{conclusion}

\FloatBarrier

\clearpage

{\small
\bibliographystyle{ieee_fullname}
\bibliography{ref}
}

\clearpage
\appendix
\label{sec:appendix}

\input{appendix}

\end{document}

%% file: intro.tex
\section{Introduction}
Visual Question Answering (VQA) has received considerable attention in the research community in recent years. The task involves predicting a textual response to a natural language question based on an image. This can be achieved by either classifying the response from a fixed set of answers or generating a free-form textual response. More recently, video question answering (VideoQA) has gained popularity as a more complex task in multimodal AI.\@ It is more challenging compared to image-based question answering because it involves reasoning about the content of objects and sequences of actions in different frames and linking them to the natural language used in the question.
\begin{figure*}[t]
\centering
\includegraphics[scale=0.25]{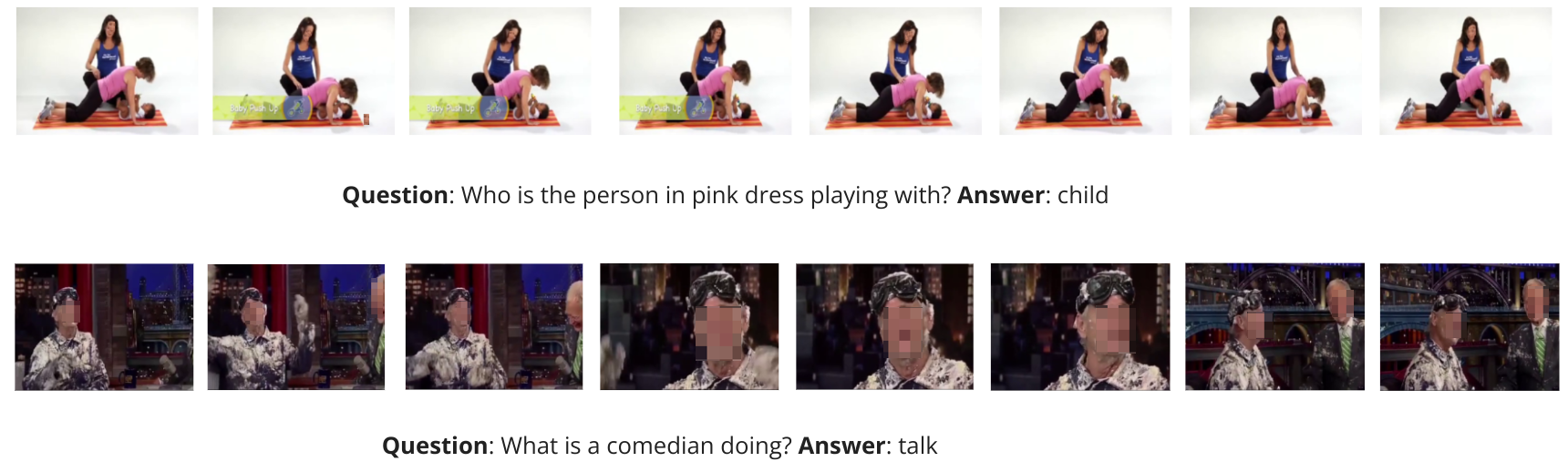}
\caption{The VideoQA task involves answering a natural language question based on a video clip.  Different from visual question answering, VideoQA requires understanding the sequence of actions/events being performed by characters in a clip and their interaction with objects to answer a question correctly. Some examples from the MSRVTT dataset \cite{DBLP:conf/mm/XuZX0Z0Z17} with the corresponding GT answer are shown.}
\label{fig:ex_msrvtt}
\end{figure*}
Consider the example shown in Figure~\ref{fig:ex_msrvtt}. Given a question - \textit{Who is the person in pink dress playing with?}, the model is tasked to infer the correct response - \textit{child}.

The Transformer model \cite{devlin2018bert} has revolutionized the field of NLP through self-supervised learning, resulting in state-of-the-art results across benchmarks  \cite{devlin2018bert, wang2018glue, sarlin2020superglue} and remarkable few-shot generalization capabilities \cite{brown2020language}. 
This has motivated similar efforts for vision-language (VL) tasks, where models like CLIP \cite{radford2021learning}, ALIGN \cite{jia2021scaling}, and Florence \cite{DBLP:journals/corr/abs-2111-11432} show promising potential for transfer learning on downstream tasks. 
Self-supervised vision-language models, such as LXMERT \cite{tan-bansal-2019-lxmert}, ViLBERT \cite{lu2019vilbert}, and VisualBERT \cite{li2019visualbert}, employ this paradigm to pre-train on a related dataset and fine-tune on downstream tasks, including VQA.\@
A line of approaches in the literature first extracts the features for each of the two modalities separately and then attempts cross-modal fusion using a simple concatenation of VL tokens \cite{li2019visualbert, DBLP:conf/iclr/SuZCLLWD20, DBLP:conf/eccv/Li0LZHZWH0WCG20} or by using cross-attention \cite{lu2019vilbert, tan-bansal-2019-lxmert}.
However, the \cotok model \cite{piergiovanni2022video} demonstrates that cross-modality interaction during the feature extraction process is a much more effective way to reason across the two modalities. 
It involves iterative learning of a set of token representations using TokenLearner \cite{ryoo2021tokenlearner} conditioned on both video and text. Such learned token representations allow the subsequent Transformer layers to process only a few tokens, resulting in more efficient models.

Representation learning is a central problem in modern machine learning. A disentangled representation is one where each feature captures information about only one salient factor of variation \cite{bengio2013representation}. Disentangled representations have superior out-of-domain (OOD) generalization \cite{Wang2022CrossDomainAD, fotiadis2021disentangled, higgins2017darla}, better interpretability \cite{adel2018discovering, higgins2018scan}, sample efficiency \cite{higgins2018scan}, and transfer learning capabilities \cite{yuan2021improving}.
To this end, we propose a different perspective on joint image-language learning, with a focus on disentangling the tokens that constitute the cross-modal feature representation.
We use a diversity enforcing loss to encourage that the feature representations are disentangled and thus representative in an economical way.  Our representations outperform the baselines in most settings and show competitive performance with state-of-the-art methods.

%% file: background.tex
\section{Background}
\noindent \textbf{TokenLearner}: TokenLearner \cite{ryoo2021tokenlearner} aims to adaptively learn a fixed set of token representations corresponding to one or more modalities, e.g. images, video, and text. The key idea is to select a series of informative combinations of spatial locations in the image/video conditioned on all modalities. More formally, let $X \in \mathbb{R}^{H\times W\times C}$ be an input image, where $H,$ $W,$ and $C$ denote the image dimensions and the no. of channels, respectively. For the $i^{th}$ token $z_i$, it learns a spatial attention map $\alpha_i(X)$ which is multiplied with the input to generate a token output $A_i(X)$,
$$ z_i = A_i(X) = \rho(X \odot \gamma(\alpha_i (X))),$$
where $\odot$ denotes Hadamard product, $\gamma(\cdot)$ denotes the broadcasting function, and $\rho(\cdot)$ denotes spatial global average pooling.\\

\noindent \textbf{Video-text iterative co-tokenization model}: The Video-text iterative co-tokenization model \cite{piergiovanni2022video} (\cotok henceforth) takes a unique approach to VideoQA by integrating interactions between video and text directly into the visual feature extraction process rather than considering them as an afterthought.
Such interactions enable better reasoning across the two modalities.
This model uses multiple streams of video at different spatio-temporal scales for multimodal representation learning. Better temporal resolution may be required to infer certain actions whereas better spatial resolution aids correct identification of objects in video frames, thus resulting in a trade-off.
This necessitates the use of multiple streams of different resolutions for superior performance. 

The \cotok model is based on a T5-style \cite{raffel2020exploring} encoder-decoder Transformer architecture. During the encoding process, the Transformer iteratively generates learned token representations in each layer by adaptive fusion of visual tokens. The input to the Transformer model is a concatenation of language features and fused visual tokens from the previous layer. Finally, the decoder generates textual output based on token representations from the last layer as the response. The model obtains initial visual and language features using an X3D model \cite{feichtenhofer2020x3d} and a T5 encoder, respectively. We use the \cotok model as the baseline for VideoQA.\@ 

%% file: model.tex
\section{Proposed Approach}

\subsection{Baseline Model}

\begin{figure}[t]
\centering
\includegraphics[width=0.95\linewidth]{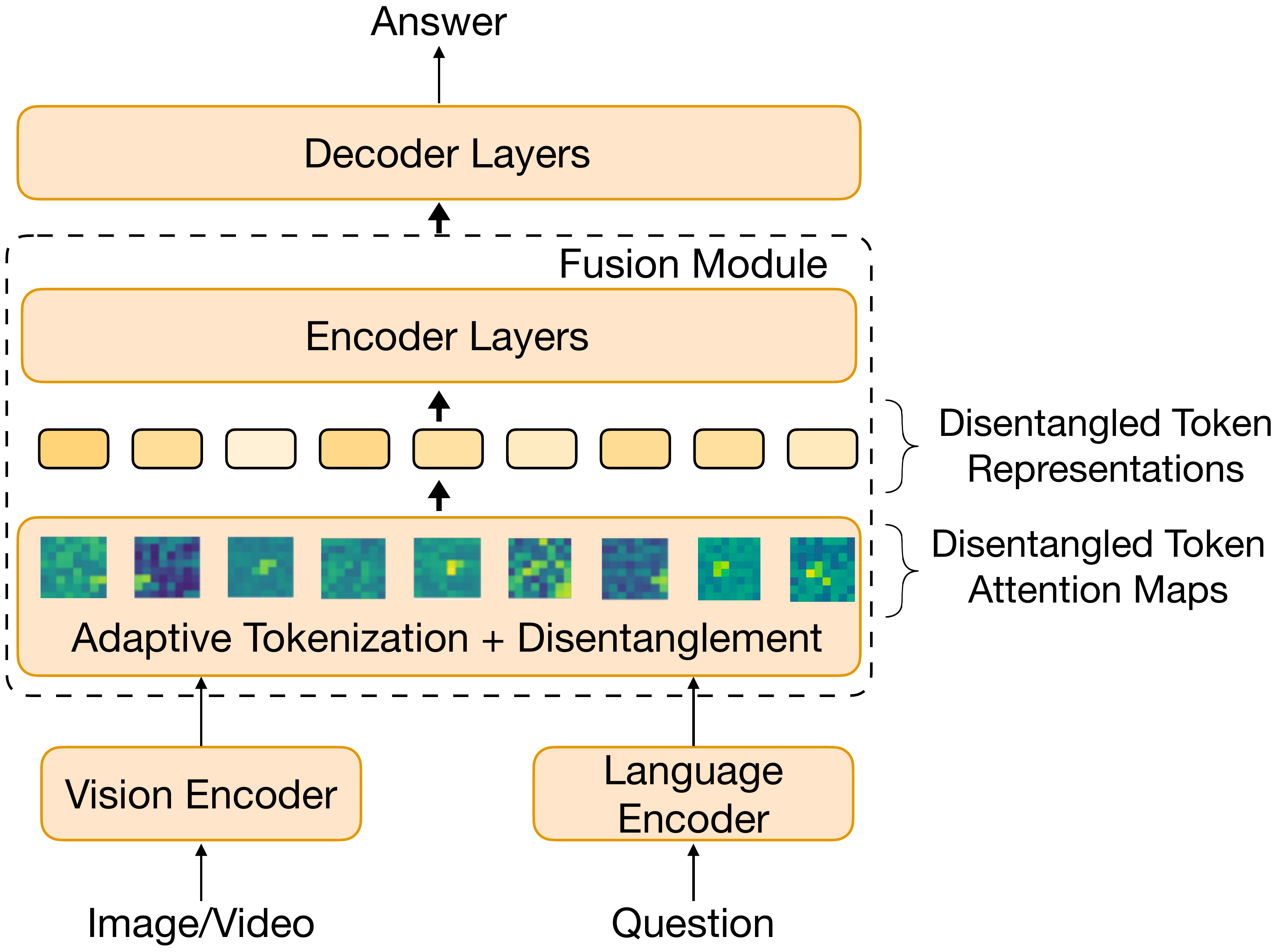}
\caption{Overall model architecture. The two modalities are fused by adaptive tokenization to generate a fixed set of \textit{disentangled} learned tokens.}
\label{fig:model}
\end{figure}

The baseline model consists of an encoder-decoder model based on T5 text encoder \cite{raffel2020exploring} and a ResNet-50 \cite{he2016deep} vision encoder. The two modalities are fused together to form a joint feature space from which a smaller set of tokens are learned. For VideoQA, we use the \cotok model as baseline. For the VQA task, we use a simplified version of the \cotok model which generates a single set of learned tokens instead of iterative tokenization.
Figure~\ref{fig:model} shows the overall model architecture.

\subsection{Disentangled representations}

To encourage the model to learn diverse token representations, we propose a new loss function - \textit{diversity loss} to be used in conjunction with the main model objective.
\begin{equation}
 \mathcal{L}_{div} = \sum_{k=1}^N\sum_{j=1, j \neq i}^M <\alpha^k_i, \alpha^k_j>^2
\end{equation}

Here, $\alpha^k_i$ denotes the spatial attention weights for the $i^{th}$ token in example $k$. $N$ and $M$ denote the number of examples and the number of fused tokens respectively. This is inspired by a similar technique to disentangle the sphere of attention of multiple agents in robotic manipulation \cite{zhang2021dair}.
The disentangled representations effectively enforce the selected representations to be orthogonal to each other. Our baseline model presented in the experiments uses identical model hyperparameters and the number of learned tokens, so as to be directly comparable to the proposed approach.

%% file: experiments.tex
\section{Experiments}

\subsection{Datasets}
\noindent We evaluate our approach on the following datasets:\\
\noindent \textbf{MSRVTT-QA} \cite{DBLP:conf/mm/XuZX0Z0Z17}: This dataset was created using MSR-VTT video descriptions dataset \cite{DBLP:conf/cvpr/XuMYR16}. It contains 10K video clips and 243K question-answer pairs.\\
\noindent \textbf{MSVD-QA} \cite{DBLP:conf/mm/XuZX0Z0Z17}: This dataset is based on Microsoft Research Video Description Corpus \cite{chen-dolan-2011-collecting}. It contains 50K questions based on 1970 video clips.\\
\noindent \textbf{IVQA} \cite{yang2021just}: It is a dataset consisting of `how-to' videos. It has 10K video clips with one question and 5 answers/clip.\\
\noindent \textbf{SNLI-VE} \cite{DBLP:journals/corr/abs-1901-06706}: The visual entailment task involves predicting whether the given statement is entailment/contradiction/neural in the context of the image.\\
\noindent \textbf{GQA} \cite{DBLP:conf/cvpr/HudsonM19}: It is a popular benchmark for visual reasoning, which was developed to address the biases of existing VQA datasets. The questions in this dataset are compositional and are grounded in Visual Genome \cite{krishna2017visual} scene graphs.\\

\subsection{Implementation Details}

For both VQA and VideoQA models, we use the Adam optimizer with a weight decay of 1e-4 and $L=12, A=12, H=768$ for the Transformer model where $L$, $A$, and $H$ denote the number of layers, the number of attention heads per layer, and the hidden size, respectively. 

The VideoQA models use 16 frames and $224\times224$ images. The two streams are $8\times224\times224$ and $16\times112\times112$.
Furthermore, in order to save compute we demonstrate competitive performance by using only one-third of the Transformer layers (and training iterations) of the original \cotok model, which allows our model to run at fewer FLOPs. For VideoQA models, pre-training is performed on a 10\% subset of the Howto69MVQA dataset \cite{yang2021just}. For VQA models, pre-training is done on the Conceptual Captions dataset \cite{changpinyo2021conceptual}. We use 8 and 16 learned tokens for VideoQA and VQA experiments, respectively.

\subsection{Results}

\noindent \textbf{VideoQA.}
We compare our approach with the baseline model both with and without pre-training (Table~\ref{tab:videoqa}). 
We observe that our approach consistently outperforms the baseline models both for both settings on MSRVTT-QA, MSVD-QA, and IVQA.\@
Table~\ref{tab:vid-sota} compares it to the state-of-the-art \cotok model, which shows the competitive performance of our model.
We note that our model has the same GFLOPS as the baseline model as the number of learned tokens is the same for both. However since the disentangled tokenization is able to select the tokens in a more economical way (i.e.\ some tokens might be `empty'), in general, fewer tokens will be needed in the end, resulting in fewer FLOPs overall.

\input{tables/videoqa_tbl}

\input{tables/videoqa_sota_tbl}

\noindent \textbf{VQA.} For the pre-training setting, our approach obtains consistent improvements on the validation set for both datasets and on the test set for SNLI-VE (Table~\ref{tab:vqa_200k}). 
Similarly, we outperform the baseline for the SNLI-VE dataset in the no pre-training setting (Table~\ref{tab:vqa_scratch}).
This shows that the proposed disentangled tokens provide better performance compared to the baseline for the same quota of tokens allocated.
Table~\ref{tab:vqa-sota} compares to the state-of-the-art models, and while our models perform well, they do not outperform particularly large models.

\input{tables/vqa_200k_tbl}

\input{tables/vqa_scratch_tbl}

\input{tables/vqa_sota_tbl}

\subsection{Visualization}

Figure ~\ref{fig:vis2} visualizes the learned disentangled representations, along with their corresponding attention maps for a VQA example. We observe that they localize their attention to much more specific areas of the image, that are vital for answering the question. Furthermore, the tokens that are selected prioritize joint visual-language representation, thereby capturing essential features from both the visual and linguistic inputs. Additional visualizations are shown in Section~\ref{sec:viz} in the Appendix.

\input{images/viz}

%% file: tables/videoqa_tbl.tex
\begin{table}[!ht]
\centering
\small
\begin{tabular}{lccc}
\toprule
\textbf{Dataset}                    &  \textbf{Pre-training}                                                                               & \textbf{Model}           & \textbf{Accuracy}           \\ \midrule
\multirow{5}{*}{MSRVTT-QA} & \multirow{2}{*}{}  & Baseline        & 31.06         \\
                           &                                                                                 & \bfseries Ours    & \textbf{31.37}         \\ \cmidrule{2-4}
                           & \multirow{2}{*}{\cmark} & Baseline        & 31.78         \\
                           &                                                                                 & \bfseries Ours    & \textbf{33.05}          \\ \midrule
                        
\multirow{5}{*}{MSVD-QA}   & \multirow{2}{*}{}  & Baseline        & 27.98         \\
                           &                                                                                 & \bfseries Ours    & \textbf{28.22}         \\ \cmidrule{2-4}
                           & \multirow{2}{*}{\cmark} & Baseline        & 28.08         \\
                           &                                                                                 & \bfseries Ours    & \textbf{30.11}         \\ \midrule
\multirow{5}{*}{IVQA}      & \multirow{2}{*}{}  & Baseline        & 9.48          \\
                           &                                                                                 & \bfseries Ours    & \textbf{9.96}         \\ \cmidrule{2-4}
                           & \multirow{2}{*}{\cmark} & Baseline        & 8.86          \\
                           &                                                                                 & \bfseries Ours    & \textbf{9.97}         \\ \bottomrule
                        
\end{tabular}
\caption{Video QA results with and without pre-training (PT) in the open vocabulary setting (validation set). The baseline is a similar capacity \cotok model.}
\label{tab:videoqa}
\end{table}

%% file: tables/videoqa_sota_tbl.tex
\begin{table}[!ht]
\centering
\small
\resizebox{\linewidth}{!}{%
\begin{tabular}{lccc}
\toprule
\bfseries Model & \bfseries MSRVTT-QA    & \bfseries MSVD-QA & \bfseries GFLOPs    \\
\midrule
Co-tok.\ ~\cite{piergiovanni2022video}  &33.7 &32.5 &67 \\          
\bfseries Ours         &33.1      & 30.1         &  41 \\
\bottomrule
\end{tabular}}
\caption{Comparison to state-of-the-art approaches for VideoQA. The comparison is done for the open-vocabulary setting. Our model FLOPs are much fewer than the prior work. We pretrain on 10\% subset of the HowTo69MVQA dataset \cite{yang2021just}, whereas \cotok pretrained on the full HowTo100M dataset \cite{miech2019howto100m}. We demonstrate competitive performance despite having a smaller model capacity.}
\label{tab:vid-sota}
\end{table}

%% file: tables/vqa_200k_tbl.tex
\begin{table}[!ht]
\centering
\small
\begin{tabular}{llllll} \toprule
        &          & \multicolumn{2}{c}{\bfseries Val.\ set} & \multicolumn{2}{c}{\bfseries Test set} \\ \cmidrule{3-6}
\bfseries Dataset & \bfseries Model    & \bfseries E.M.      & \bfseries F1          & \bfseries E.M.       & \bfseries F1          \\ \midrule
SNLI-VE & Baseline & 76.70          & 76.70       & 76.59           & 76.59        \\
        & \bfseries Ours     & \textbf{78.06}          & \textbf{78.06}       & \textbf{77.36}          & \textbf{77.37} \\ \midrule    
GQA     & Baseline & 73.48          & 73.56       & \textbf{73.5}          & \textbf{73.57}        \\
        & \bfseries Ours     & \textbf{75.02}          & \textbf{75.11}       & 75.01          & 75.1       \\ \bottomrule
\end{tabular}
\caption{VQA results in the pre-training setting.}
\label{tab:vqa_200k}
\end{table}

%% file: tables/vqa_scratch_tbl.tex
\begin{table}[!ht]
\centering
\small
\begin{tabular}{llllll} \toprule
        &          & \multicolumn{2}{c}{\bfseries Val.\ set} & \multicolumn{2}{c}{\bfseries Test set} \\ \cmidrule{3-6}
\bfseries Dataset & \bfseries Model    & \bfseries E.M.       & \bfseries F1          & \bfseries E.M.       & \bfseries F1          \\ \midrule
SNLI-VE & Baseline & 73.08          & 73.08       & 72.5           & 72.5        \\
        & \bfseries Ours     & \textbf{73.15}          & \textbf{73.15}       & \textbf{72.69}          & \textbf{72.69} \\ \midrule
GQA     & Baseline & \textbf{68.08}          & \textbf{68.13}       & \textbf{68.14}          & \textbf{68.2}        \\
        & \bfseries Ours     & 67.98          & 68.02       & 67.98          & 68.02       \\ \bottomrule      
\end{tabular}
\caption{VQA results in the no pre-training setting.}
\label{tab:vqa_scratch}
\end{table}

%% file: tables/vqa_sota_tbl.tex
\begin{table}[!ht]
\centering
\begin{tabular}{l|ll}
\toprule
\bfseries Model & \bfseries GQA    & \bfseries SNLI-VE    \\
\midrule
SimVLM (Huge)~\cite{wang2022simvlm}  & -- & \bfseries 86.32 \\ 
\midrule
UNITER~\cite{chen2020uniter}  & -- & 79.38 \\ 
VinVL~\cite{zhang2021vinvl}        & 65.05 & --   \\
LXMERT~\cite{tan-bansal-2019-lxmert}         & 60.0 & -- \\
\midrule         
\bfseries Ours         & \textbf{76.79}      & 80.15              \\
\bottomrule
\end{tabular}
\caption{Comparison to state-of-the-art approaches (test-dev set for GQA and test set for SNLI-VE). Missing values are denoted by --. Our model uses a 3M dataset for pre-training and has about 300M parameters. Approaches which use much larger data or model are shown in the top section.}
\label{tab:vqa-sota}
\end{table}

%% file: images/viz.tex
\begin{figure}[htbp]
\centering
\begin{subfigure}[c]{\linewidth}
\centering
\includegraphics[width=0.7\linewidth]{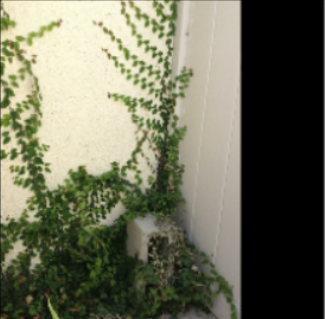}
\caption{Question Image}
\end{subfigure}
\vskip\baselineskip
\begin{subfigure}[b]{0.475\textwidth}
\centering
\hspace*{\fill}%
\includegraphics[width=0.4\linewidth]{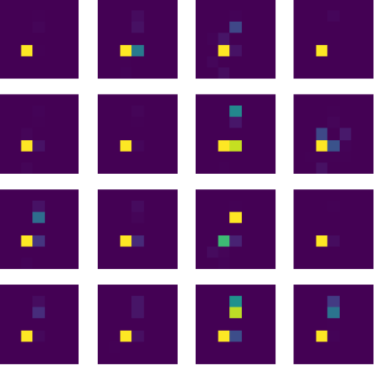} \hfill
\includegraphics[width=0.4\linewidth]{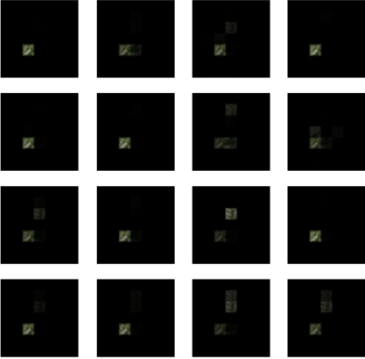}
\hspace*{\fill}%
\caption{Token visualization (Baseline)}
 \end{subfigure}

\begin{subfigure}[b]{0.475\textwidth}
\centering
\hspace*{\fill}%
\includegraphics[width=0.4\linewidth]{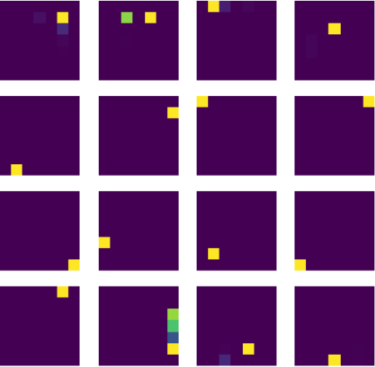} \hfill
\includegraphics[width=0.4\linewidth]{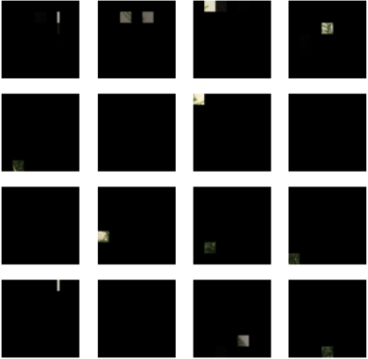}
\hspace*{\fill}%
\caption{Token visualization (Ours)}
 \end{subfigure}
\caption{(a) \textit{Question}: what kind of climbing vine or plant is this?
\textit{Baseline}: tombppry,
\textit{Ours}: \underline{ivy},
\textit{Ground truth answers} = [`fern', `grape', `vine', \underline{`ivy'}, `unanswerable', `creeping fig', `unanswerable', `unanswerable', \underline{`ivy'}, `green']; \textit{Bottom left}: Weights assigned to each image patch for every token, lighter shades like yellow correspond to higher weights; \textit{Bottom right}: Token attention masks grounded to the input image.}
\label{fig:vis2}
\end{figure}

%% file: conclusion.tex
\section{Conclusion}
In this work, we propose learning disentangled representations for the learned tokens in Transformer models for VQA and VideoQA tasks. This simple-yet-effective approach leads to a performance boost in a majority of training settings across datasets. Future work will involve benchmarking this approach with higher capacity models and more pre-training for improved performance. Another promising future direction is to utilize the learned token representations for related downstream tasks.

%% file: appendix.tex
\setcounter{table}{0}
\renewcommand\thetable{\Alph{section}.\arabic{table}}
\setcounter{figure}{0}
\renewcommand\thefigure{\Alph{section}.\arabic{figure}}

\section{Visualizations}\label{sec:viz}

\FloatBarrier

\begin{figure}[htbp]
\centering
\begin{subfigure}[c]{\linewidth}
\centering
\includegraphics[width=0.7\linewidth]{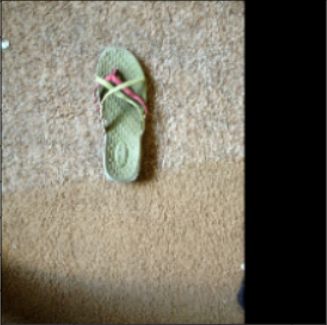}
\caption{Question Image}
\end{subfigure}
\vskip\baselineskip
\begin{subfigure}[b]{0.475\textwidth}
\centering
\hspace*{\fill}%
\includegraphics[width=0.4\linewidth]{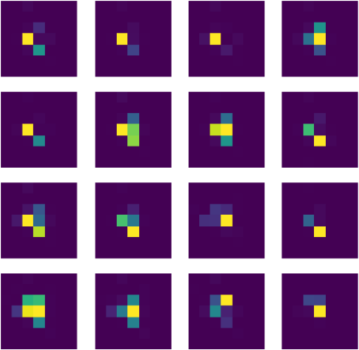} \hfill
\includegraphics[width=0.4\linewidth]{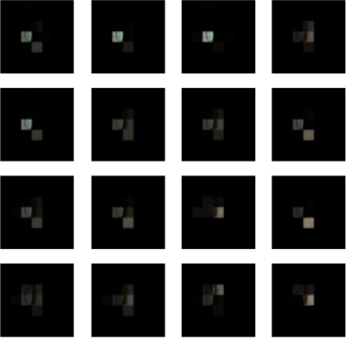}
\hspace*{\fill}%
\caption{Token visualization (Baseline)}
\end{subfigure}
\vskip\baselineskip
\begin{subfigure}[b]{0.475\textwidth}
\centering
\hspace*{\fill}%
\includegraphics[width=0.4\linewidth]{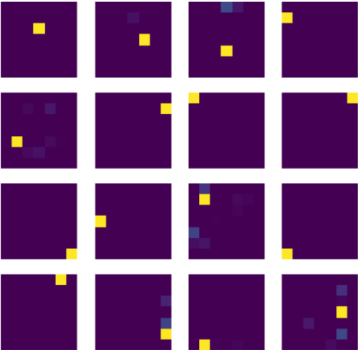} \hfill
\includegraphics[width=0.4\linewidth]{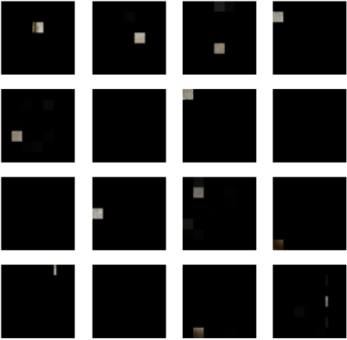}
\hspace*{\fill}%
\caption{Token visualization (Ours)}
\end{subfigure}
\caption{\textit{Question}: what color are the straps on the sandal?
\textit{Baseline}: pink sugar,
\textit{Ours}: \underline{pink green},
\textit{Ground truth answers} = [\underline{`pink green'}, \underline{`pink green'}, `slipper', \underline{`pink green'}, `green pink', `pink mint green', `green pink', \underline{`pink green'}, `green pink', `green']; \textit{Bottom left}: Weights assigned to each image patch for every token, lighter shades like yellow correspond to higher weights; \textit{Bottom right}: Token attention masks grounded to the input image.}
\label{fig:vis_2}
\end{figure}

\begin{figure}[htbp]
\centering
\begin{subfigure}[c]{\linewidth}
\centering
\includegraphics[width=0.7\linewidth]{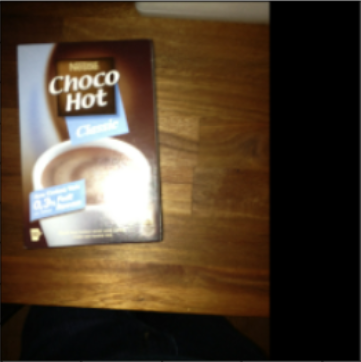}
\caption{Question Image}
\end{subfigure}
\vskip\baselineskip
\begin{subfigure}[b]{0.475\textwidth}
\centering
\hspace*{\fill}%
\includegraphics[width=0.4\linewidth]{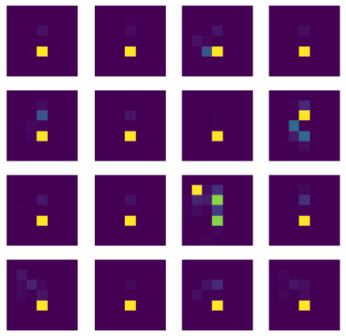} \hfill
\includegraphics[width=0.4\linewidth]{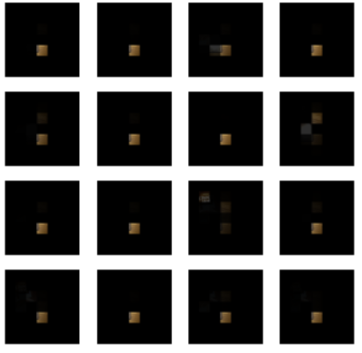}
\hspace*{\fill}%
\caption{Token visualization (Baseline)}
\end{subfigure}
\vskip\baselineskip
\begin{subfigure}[b]{0.475\textwidth}
\centering
\hspace*{\fill}%
\includegraphics[width=0.4\linewidth]{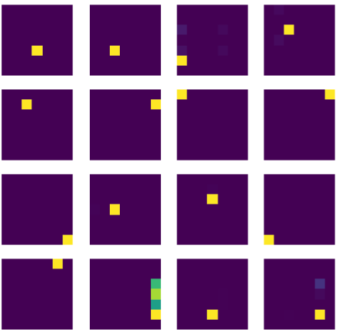} \hfill
\includegraphics[width=0.4\linewidth]{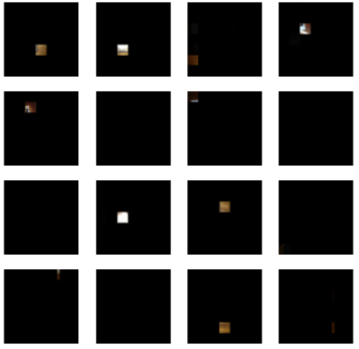}
\hspace*{\fill}%
\caption{Token visualization (Ours)}
\end{subfigure}
\caption{\textit{(a) Question}: what is this?
\textit{Baseline}: hot baby dar1,
\textit{Ours}: \underline{hot chocolate mix},
\textit{Ground truth answers} = ['cocoa mix', 'hot cocoa', \underline{'hot chocolate mix'}, 'choco hot', 'choco hot', 'choco hot', 'hot cocoa', 'hot chocolate', 'hot chocolate', 'hot cocoa']; \textit{Bottom left}: Weights assigned to each image patch for every token, lighter shades like yellow correspond to higher weights; \textit{Bottom right}: Token attention masks grounded to the input image.}
\label{fig:vis_3}
\end{figure}